\documentclass{llncs}
\usepackage[colorinlistoftodos]{todonotes}

\usepackage[pagebackref=true]{hyperref}

\usepackage{amssymb}
\usepackage{bm}
\usepackage{subfigure}
\usepackage{amsmath,graphicx,dsfont,color}
\usepackage{amsfonts}
\usepackage{amssymb}
\usepackage{array}
\usepackage{algorithm}
\usepackage{algorithmic}
\usepackage{placeins}
\usepackage{lipsum}

\newcommand{\squeezeup}{\vspace{-2.5mm}}
\newcommand{\squeezeupsmall}{\vspace{-1mm}}

\newcommand{\nsize}{n}
\newcommand{\node}{v}
\newcommand{\realset}{\mathbb{R}}
\newcommand{\I}{\mathbf{I}}
\newcommand{\graph}{\mathcal{G}}
\newcommand{\mst}{\mathcal{MST}}
\newcommand{\msf}{\mathcal{F}}
\newcommand{\nodes}{\mathcal{V}}
\newcommand{\edges}{\mathcal{E}}
\newcommand{\partition}{\mathsf{\pi}}
\newcommand{\region}{\mathsf{R}}
\newcommand{\hierarchy}{\mathcal{H}}

\newcommand{\fultra}{\bm{\thUCM}}
\newcommand{\W}{\mathbf{W}}

\newcommand{\probamap}{\mathrm{PM}}
\newcommand{\thUCM}{\lambda}

\newcommand{\tree}{$T$} 

\newcommand{\nummarkers}{N}

\begin{document}
\title{Prior-based Hierarchical Segmentation Highlighting Structures of Interest}
\author{Amin Fehri, Santiago Velasco-Forero and Fernand Meyer\\
\texttt{\{amin.fehri,santiago.velasco,fernand.meyer@mines-paristech.fr\}}}
\institute{MINES ParisTech, PSL Research University, Center for Mathematical Morphology}
\maketitle
\begin{abstract}
Image segmentation is the process of partitioning an image into a set of meaningful regions according to some criteria. Hierarchical segmentation has emerged as a major trend in this regard as it favors the emergence of important regions at different scales. On the other hand, many methods allow us to have prior information on the position of structures of interest in the images. In this paper, we present a versatile hierarchical segmentation method that takes into account any prior spatial information and outputs a hierarchical segmentation that emphasizes the contours or regions of interest while preserving the important structures in the image. Several applications are presented that illustrate the method versatility and efficiency.
\end{abstract}

\section{Introduction}

In this paper, we propose a method to take advantage of any prior spatial information previously obtained on an image to get a hierarchical segmentation of this image that emphasizes its regions of interest, allowing us to get more details in the designated regions of interest of an image while still preserving its strong structural information.

Potential applications are numerous. When having a limited storage capacity (for very large images for example), this would allow us to keep details in the regions of interest as a priority. 
Similarly, in situations of transmission with limited bandwidth, one could first transmit the important information of the image: the details of the face for a video-call, the pitch and the players for a soccer game and so on. One could also use such a tool as a preprocessing one, for example to focus on an individual from one camera view to the next one in video surveillance tasks. Finally, from an artistic point of view, the result is interesting and similar to a combination of focus and cartoon effects. Some of these examples are illustrated in this paper.

Image segmentation has been shown to be inherently a multi-scale problem \cite{guigues2006scale}. That is why hierarchical segmentation has become the major trend in image segmentation and most top-performance segmentation techniques \cite{arbelaez2011contour}\cite{pont2015multiscale}\cite{Ren13}\cite{meyer15} fall into this category: hierarchical segmentation does not output a single partition of the image pixels into sets but instead a single multi-scale structure that aims at capturing relevant objects at all scales. Researches on this topic are still vivid as differential area profiles \cite{ouzounis12}, robust segmentation of high-dimensional data \cite{gueguen2013local} as well as theoretical aspects regarding the concept of partition lattice \cite{Serra12}\cite{Ronse2013} and optimal partition in a hierarchy \cite{kiran2013ground}\cite{Ravi2013global}\cite{xu2016hierarchical}.
Our goal in this paper is to develop a hierarchical segmentation algorithm that focuses on certain predetermined zones of the image. The hierarchical aspect also allows us, for tasks previously described, to very simply tune the level of details wanted depending on the application. 

Furthermore, our algorithm is very versatile, as the spatial prior information that it uses can be obtained by any of the numerous learning-based approaches proposed over the last decades to roughly localize objects \cite{oquab15} \cite{Lampert08} \cite{Sermanet13}. In this regard, our work joins an important research point that consists in designing approaches to incorporate prior knowledge in the segmentation, as shape prior on level sets \cite{chan2005level}, star-shape prior by graph-cut \cite{veksler2008star}, use of a shape prior hierarchical characterization obtained with deep learning \cite{chen2013deep}, or related work making use of stochastic watershed to perform targeted image segmentation \cite{Malmberg17}.  

The remainder of the paper is organized as follows. Part \ref{sec:hierarchies} explains how we construct and use graph-based hierarchical segmentation. Then part \ref{sec:ourmethod} specifies how we use prior information on the image to obtain hierarchies with regionalized fineness. Several examples of applications of this method are described in part \ref{sec:examples}. Finally, conclusions and perspectives are presented in part \ref{sec:conclusion}.        

\section{Hierarchies and partitions}
\label{sec:hierarchies}
\squeezeup
\squeezeup
\subsection{Graph-based hierarchical segmentation} 

Obtaining a suitable segmentation directly from an image is very difficult. This is why it is often make use of hierarchies to organize and propose interesting contours by valuating them. In this section, we remind the reader how to construct and use graph-based hierarchical segmentation. 

For each image, let us suppose that a fine partition is produced by an initial segmentation (for instance a set of superpixels \cite{Achanta12,machairas2015waterpixels} or the basins produced by a classical watershed algorithm \cite{meyer1990morphological}) and contains all contours making sense in the image. We define a dissimilarity measure between adjacent tiles of this fine partition. 
One can then see the image as a graph, the \textit{region adjacency graph} (RAG), in which each node represents a tile of the partition; an edge links two nodes if the corresponding regions are neighbors in the image; the weight of the edge is equal to the dissimilarity between these regions. Working on the RAG is much more efficient than working on the image, as there are far less nodes in the RAG than there are pixels in the image.

Formally, we denote this graph $\graph=(\nodes,\edges,\W)$, where $\nodes$ corresponds to the image domain or set of pixels/fine regions, $\edges \subset \nodes \times \nodes$ is the set of edges linking neighbour regions,   
$\W: \edges \to \realset^{+}$ is the dissimilarity measure usually based on local gradient information (or color or texture), for instance $\W(i,j) \propto |\I(\node_i)-\I(\node_j)|$ with $\I:\nodes \to \realset$ representing the image intensity. 

The edge linking the nodes $p$ and $q$ is designated by $e_{pq}$ . A path is a sequence of nodes and edges: for example the path linking the nodes $p$ and $s$ is the set $\{p, e_{pt} , t, e_{ts}, s\}$. A \textit{connected subgraph} is a subgraph where each pair of nodes is connected by a path. A \textit{cycle} is a path whose extremities coincide. A \textit{tree} is a connected graph without cycle. A \textit{spanning tree} is a tree containing all nodes. A \textit{minimum spanning tree} (MST) $\mst(\graph)$ of a graph $\graph$ is a spanning tree with minimal possible weight, obtained for example using the Boruvka algorithm (the weight of a tree being equal to the sum of the weights of its edges). A \textit{forest} is a collection of trees.

A \emph{partition} $\partition$ of a set $\nodes$ is a collection of subsets of $\nodes$, such that the whole set $\nodes$ is the disjoint union of the subsets in the partition, i.e., $\partition=\{\region_1,\region_2,\ldots,\region_k\}$, such that $\forall i, \region_i \subseteq \nodes \ $; $ \forall i\neq j,  \region_i \cap\region_j = \emptyset$ ; $\bigcup_{i}^{k}\region_i = \nodes$. 

Cutting all edges of the $\mst(\graph)$ having a valuation superior to a threshold $\thUCM$ leads to a minimum spanning forest (MSF) $\msf(\graph)$, i.e. to a partition of the graph. Note that the obtained partition is the same that one would have obtain by cutting edges superior to $\thUCM$ directly on $\graph$ \cite{najman13}. Since working on the $\mst(\graph)$ is less costly and provides similar results regarding graph-based segmentation, we work only with the $\mst(\graph)$ in the sequel.

\squeezeup
\squeezeup
\begin{figure}
\begin{center}
\includegraphics[width=.55 \columnwidth]{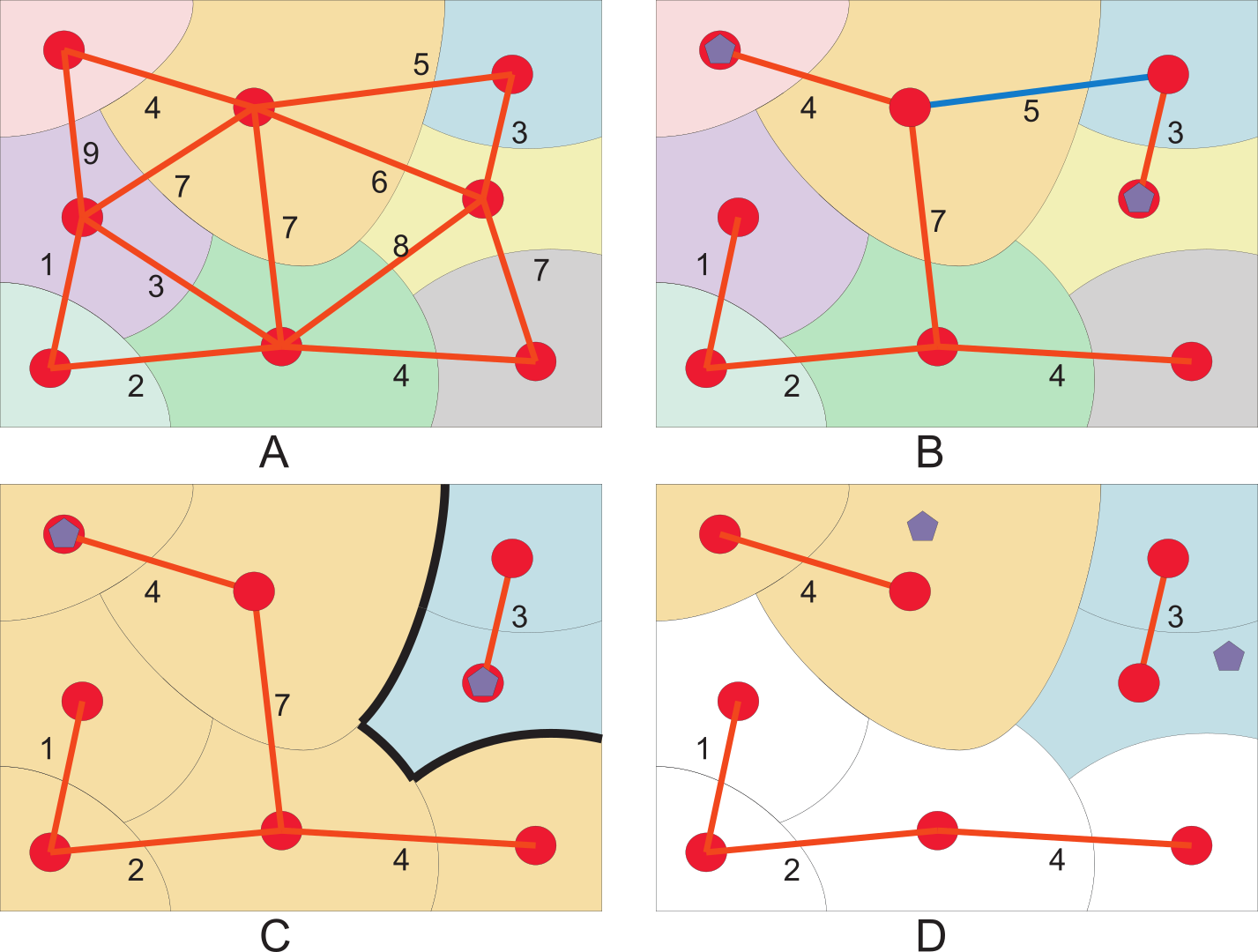}
\squeezeup
\caption{\textbf{A}: a partition represented by an edge-weighed graph; \textbf{B}: a minimum spanning tree of the graph, with 2 markers in blue: the highlighted edge in blue is the highest edge on the path linking the two markers ; \textbf{C}: the segmentation obtained when cutting this edge; \textbf{D} blue and orange domain are the domains of variation of the two markers generating the same segmentation.}\label{Fig:MST}
\end{center}
\squeezeup
\end{figure}

\squeezeup
\squeezeup

So cutting edges by decreasing valuations gives an \emph{indexed hierarchy of partitions} $(\hierarchy,\fultra)$, with $\hierarchy$ a \emph{hierarchy of partitions} i.e. a chain of nested partitions $\hierarchy=\{\partition_0, \partition_1,\ldots, \partition_\nsize| \forall j,k, \quad 0 \quad \leq j\leq k\leq \nsize \Rightarrow \partition_j \sqsubseteq \partition_k\}$, with $\partition_\nsize$ the single-region partition and $\partition_0$ the finest partition on the image, and $\fultra: \hierarchy \to \realset^+$  being a stratification index verifying $\fultra(\partition) < \fultra(\partition')$ for two nested partitions $\partition \subset \partition'$. This increasing map allows us to value each contour according to the level of the hierarchy for which it disappears: this is the \emph{saliency} of the contour, and we consider that the higher the saliency, the stronger the contour.
For a given hierarchy, the image in which each contour takes as value its saliency is called \textit{Ultrametric Contour Map} (UCM)\cite{arbelaez2011contour}. Representing a hierarchy by its UCM is an easy way to get an idea of its effect because thresholding an UCM always provides a set of closed curves and so a partition. In this paper, for better visibility, we represent UCM with inverted contrast.

To get a partition for a given hierarchy, there are several possibilities: 
\squeezeupsmall
\begin{itemize}
\item simply thresholding the highest saliency values,
\item marking some nodes as important ones and then computing a partition accordingly, which is known as \textit{marker-based segmentation},
\item smartly editing the graph by finding the partition that minimizes an energetic function.
\end{itemize}
\squeezeupsmall
In a complementary approach, we argue that the quality of the obtained partitions highly depends on the hierarchy that we use, and thus that changing the dissimilarity can lead to more suitable partitions. Indeed, if the dissimilarity reflects only a local contrast as in the hierarchy issued by the RAG, the most salient regions in the image are the small contrasted ones.  So instead of departing from a simple and rough dissimilarity such as contrast and then use an sophisticated technique to get a good partition out of it, one can also try to obtain a more informative dissimilarity adapted to the content of the image such that the simplest methods are sufficient to compute interesting partitions. This way, the aforementioned techniques lead to segmentations better suited for further exploitation. 
How can we construct more pertinent and informative dissimilarities?

\squeezeup
\squeezeup
\subsection{Stochastic watershed hierarchies}
\label{ssec:stochastic}
The stochastic watershed (SWS), introduced in \cite{angulo2007stochastic} on a simulation basis and extended with a graph-based approach in \cite{meyer15}, is a versatile tool to construct hierarchies. The seminal idea is to operate multiple times marker-based segmentation with random markers and valuate each edge of the $\mst$ by its frequency of appearance in the resulting segmentations.

Indeed, by spreading markers on the RAG $\graph$, one can construct a segmentation as a MSF $\msf(\graph)$ in which each tree takes root in a marked node. Marker-based segmentation directly on the $\mst$ is possible: one must then cut, for each pair of markers, the highest edge on the path linking them. Furthermore, there is a domain of variation in which each marker can move while still leading to the same final segmentation. More details are provided in Figure \ref{Fig:MST}. 

Let us then consider on the $\mst$ an edge $e_{st}$ of weight $\omega_{st}$ and compute its probability to be cut. We cut all edges of the $\mst$ with a weight superior or equal to $\omega_{st}$, producing two trees $\tree_{s}$ and $\tree_{t}$ of roots $s$ and $t$. If at least one marker falls within the domain $\region_{s}$ of $\tree_{s}$ nodes and at least one marker falls within the domain $\region_{t}$ of $\tree_{t}$ nodes, then $e_{st}$ will be cut in the final segmentation.

Let denote $\mu(\region)$ the number of random markers falling in a region $\region$. We want to attribute to $e_{st}$ the following probability value:
\begin{equation} \label{Proba}
\begin{split}
\mathbb{P}[(\mu(\region_{s}) \geq 1) \land (\mu(\region_{t}) \geq 1)] & = 1-\mathbb{P}[(\mu(\region_{s}) = 0) \lor (\mu(\region_{t}) = 0)] \\
& = 1-\mathbb{P}(\mu(\region_{s}) = 0)-\mathbb{P}(\mu(\region_{t}) = 0)+\mathbb{P}(\mu(\region_{s} \cup \region_{t}) = 0) \\
\end{split}
\end{equation}

If markers are spread following a Poisson distribution, then for a region $R$:
\begin{equation} \label{Poisson}
\mathbb{P}(\mu(R) = 0)=\exp^{-\Lambda(R)}, 
\end{equation}
With $\Lambda(R)$ being the expected value (mean value) of the number of markers falling in $R$. The probability thus becomes:
\begin{equation} \label{newProba}
\mathbb{P}(\mu(\region_{s}) \geq 1 \land \mu(\region_{t}) \geq 1) = 1-\exp^{-\Lambda(\region_{s})}-\exp^{-\Lambda(\region_{t})}+\exp^{-\Lambda(\region_{s} \cup \region_{t})}
\end{equation}

When the Poisson distribution has an homogeneous density $\lambda$:
\begin{equation} \label{uniform}
\Lambda(R) = \texttt{area}(R) \lambda, 
\end{equation}
When the Poisson distribution has a non-uniform density $\lambda$:
\begin{equation} \label{nonuniform}
\Lambda(R)=\int_{(x,y) \in R} \lambda(x,y) \, \mathrm dx \mathrm dy
\end{equation}

The output of the SWS algorithm thus depends on the departure $\mst$ (structure and edges valuations) and of the probabilistic law governing the markers distribution. Furthermore, SWS hierarchies can be chained, leading to a wide exploratory space that can be used in a segmentation workflow \cite{fehri16}.

Because of its versatility and good performance, SWS represents a good departure algorithm to modify in order to inject prior information. Indeed, when having a prior information about the image, is it possible to use it in order to have more details in some parts rather than others?

%

\section{Hierarchies highlighthing structures of interest using prior information}
\label{sec:ourmethod}

\subsection{Hierarchy with Regionalized Fineness (HRF)}
\label{ssec:HRF}

In the original SWS, a uniform distribution of markers is used (whatever size or form they may have). In order to have stronger contours in a specific region of the image, we adapt the model so that more markers are spread in this region. 

Let $E$ be an object or class of interest, for example $E = {\text{``face of a person"}}$, and $\I$ be the studied image. We denote by $\theta_{E}$ the probability density function (PDF) associated with $E$ obtained separately, and defined on the domain $D$ of $\I$, and by $\probamap(\I,\theta_{E})$ the probabilistic map associated, in which each pixel $p(x,y)$ of $\I$ takes as value $\theta_{E}(x,y)$ its probability to be part of $E$. Given such an information on the position of an event in an image, we obtain a hierarchical segmentation focused on this region by modulating the distribution of markers. 


If $\lambda$ is a density defined on $D$ to distribute markers (uniform or not), we set $\theta_{E} \lambda$ as a new density, thus favoring the emergence of contours within the regions of interest. 

Considering a region $\region$ of the image, the mean number of markers falling within $\region$ is then:
\begin{equation} \label{newnonuniform}
\Lambda_{E}(\region)=\int_{(x,y) \in \region} \theta_{E}(x,y) \lambda(x,y) \, \mathrm dx \mathrm dy
\end{equation}

Note that if we want $\nummarkers$ markers to fall in average within the domain $D$, we work with a slightly modified density:
\begin{equation} \label{eqNbMarkers}
\hat{\lambda}=\frac{\nummarkers}{\mu(D)} \lambda  
\end{equation} 

Furthermore, this approach can be easily extended to the case where we want to take advantage of information from multiple sources. Indeed, if $\theta_{E_1}$ and $\theta_{E_2}$ are the PDF associated with two events $E_1$ and $E_2$, we can combine those two sources by using as a new density $(\theta_{E_1}+\theta_{E_2})\lambda$.

\subsection{Methodology}
We present here the steps to compute a HRF for an event $E$ given a probabilistic map $\probamap(\I,\theta_{E})$ providing spatial prior information on an image $\I$:
\begin{itemize}
\item compute a fine partition $\pi_{0}$ of the image, define a dissimilarity measure between adjacent regions and compute the RAG $\graph$, and then the $\mst(\graph)$ to easily work with graphs,
\item compute a probabilistic map $\pi_{\mu}=\pi_{\mu}(\pi_{0},\probamap(\I,\theta_{E}))$ with each region of the fine partition $\pi_{0}$ taking as a new value the mean value of $\probamap(\I,\theta_{E}))$ in this region, 
\item compute new values of edges by a bottom-up approach as described in section \ref{ssec:HRF}, where for each region $\region_{i}$ of $\pi_{0}$, $\Lambda(\region_{i})$ corresponds to the mean value taken by pixels of the region $\region_{i}$ in $\pi_{\mu}$. Note that this approach allows a highly efficient implementation using dynamic programming on graphs.
\end{itemize}


\subsection{Modulating the HRF depending on the couple of regions considered}
\label{ssec:varHRF}

If we want to favor certain contours to the detriment of others, we can modulate the density of markers in each region by taking into account the strength of the contour separating them but also the relative position of both regions.

We use the same example and notations as in section \ref{ssec:HRF}, and thus want to modulate the distribution of markers relatively to $\region_{s}$, $\region_{t}$ and their frontier. 
For example, to stress the strength of the gradient separating both regions we can locally spread markers following the distribution $\chi(\region_{s},\region_{t}) \lambda$, with $\chi(\region_{s},\region_{t})=\omega_{st}$. This corresponds to the classical volume-based SWS, which allows to obtain a hierarchy that takes into account both surfaces of regions and contrast between them.

To go further, one can use any prior information in a similar way. Indeed, while using prior information to influence the output of the segmentation workflow, one might also want to choose whether the relevant information to emphasize in resulting segmentations is the foreground, the background or the transitions between them. 

For example, having more details in the transition regions between background and foreground allows us to have more precision where the limit between foreground and background is actually unclear. As a matter of fact, the prior information often only provides rough positions of the foreground object with blurry contours, and such a process would allow to get precise contours of this object from the image.

Let us consider this case and define for each couple of regions $(\region_{s},\region_{t})$ a suitable $\chi(\region_{s},\region_{t})$. We then want $\chi(\region_{s},\region_{t})$ to be low if $\region_{s}$ and $\region_{t}$ both are in the background or the foreground, and high if $\region_{s}$ is the background and $\region_{t}$ in the foreground (or the opposite). We use:
\begin{equation}\label{eqNbMarkers2}
\left\{
\begin{array}{ll}
	\tilde{\lambda}= \chi \lambda \\
	\chi(\region_{s},\region_{t}) = \frac{\max(m(\region_{s}),m(\region_{t}))(1-\min(m(\region_{s}),m(\region_{t})))}{0.01+\sigma(\region_{s})\sigma(\region_{t})},
\end{array}
\right.
\end{equation}
$m(\region)$ (resp. $\sigma(\region)$) being the normalized mean (resp. normalized standard deviation) of pixels values in the region $\region$ of $\probamap(\I)$.
Thus the number of markers spread will be higher when the contrast between adjacent regions is high (numerator term) and when these regions are coherent (denominator term).

Then for each edge, its new probability to be cut is :
\begin{align} \label{newProba2}
\begin{split}
\mathbb{P}[(\mu(\region_{p}) \geq 1) \land (\mu(\region_{q}) \geq 1)] &=  1-\exp^{-\chi(\region_{s},\region_{t})\Lambda(\region_{p})}-\exp^{-\chi(\region_{s},\region_{t}) \Lambda(\region_{q})} \\ 
& + \exp^{-\chi(\region_{s},\region_{t}) \Lambda(\region_{p} \cup \region_{q})}
\end{split}
\end{align}

In the spirit of \cite{Chen16}, this mechanism provides us with a way to ``realign" the hierarchy with respect to the relevant prior information to get more details where the information is blurry. Similar adaptations can be thought of to emphasize details of background or foreground regions.

In the following, we illustrate the methodology exposed with some applications.

\begin{figure}[H]
\squeezeup
\squeezeup
\squeezeup
\begin{center}
\subfigure[]{\includegraphics[width=.242\columnwidth]{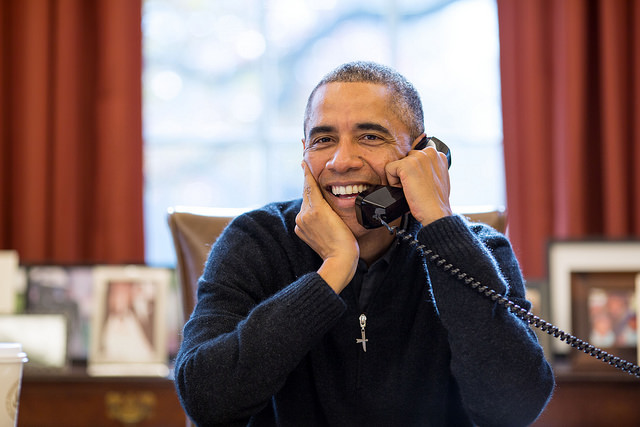}}
\subfigure[]{\includegraphics[width=.242\columnwidth]{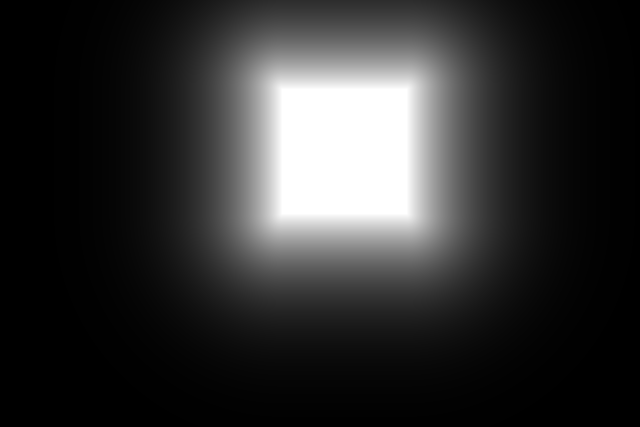}}
\subfigure[]{\includegraphics[width=.242\columnwidth]{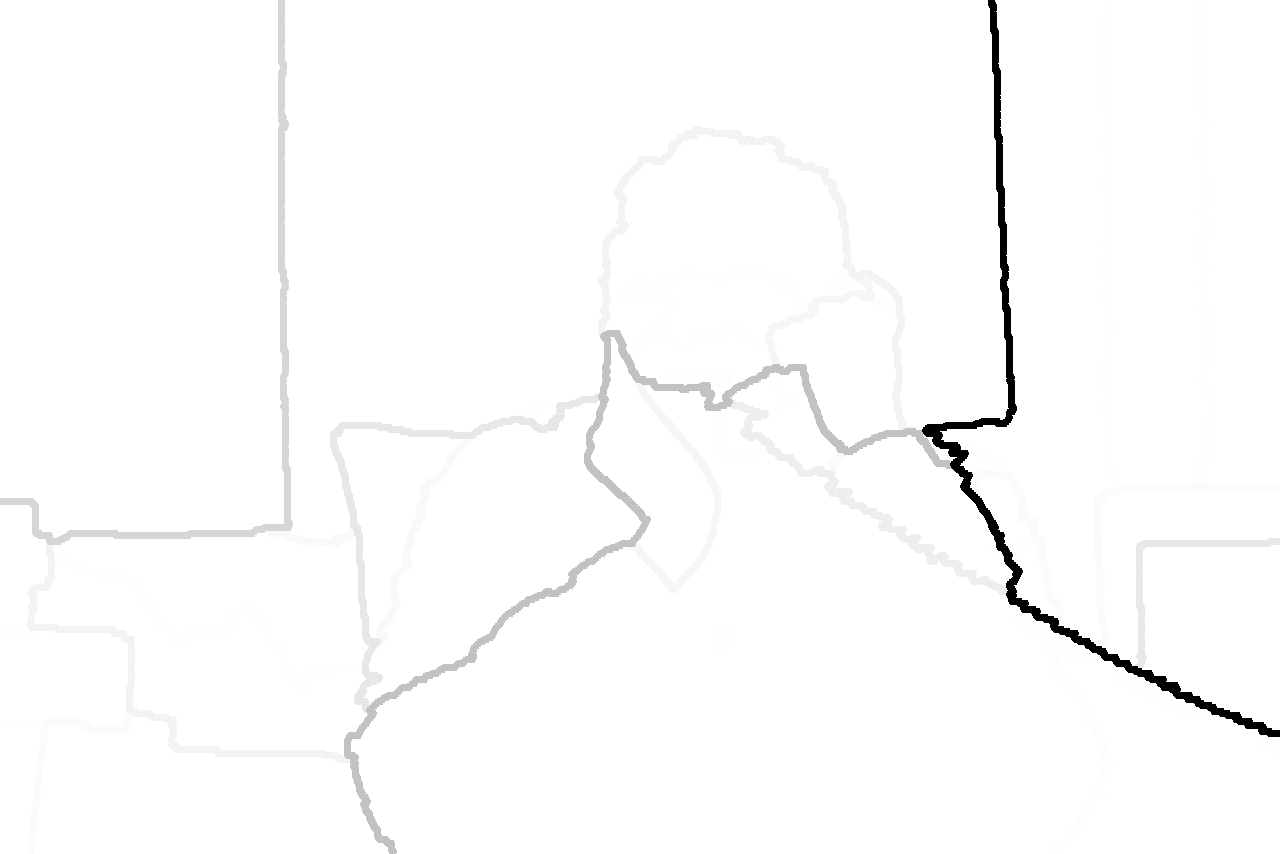}}
\subfigure[]{\includegraphics[width=.242\columnwidth]{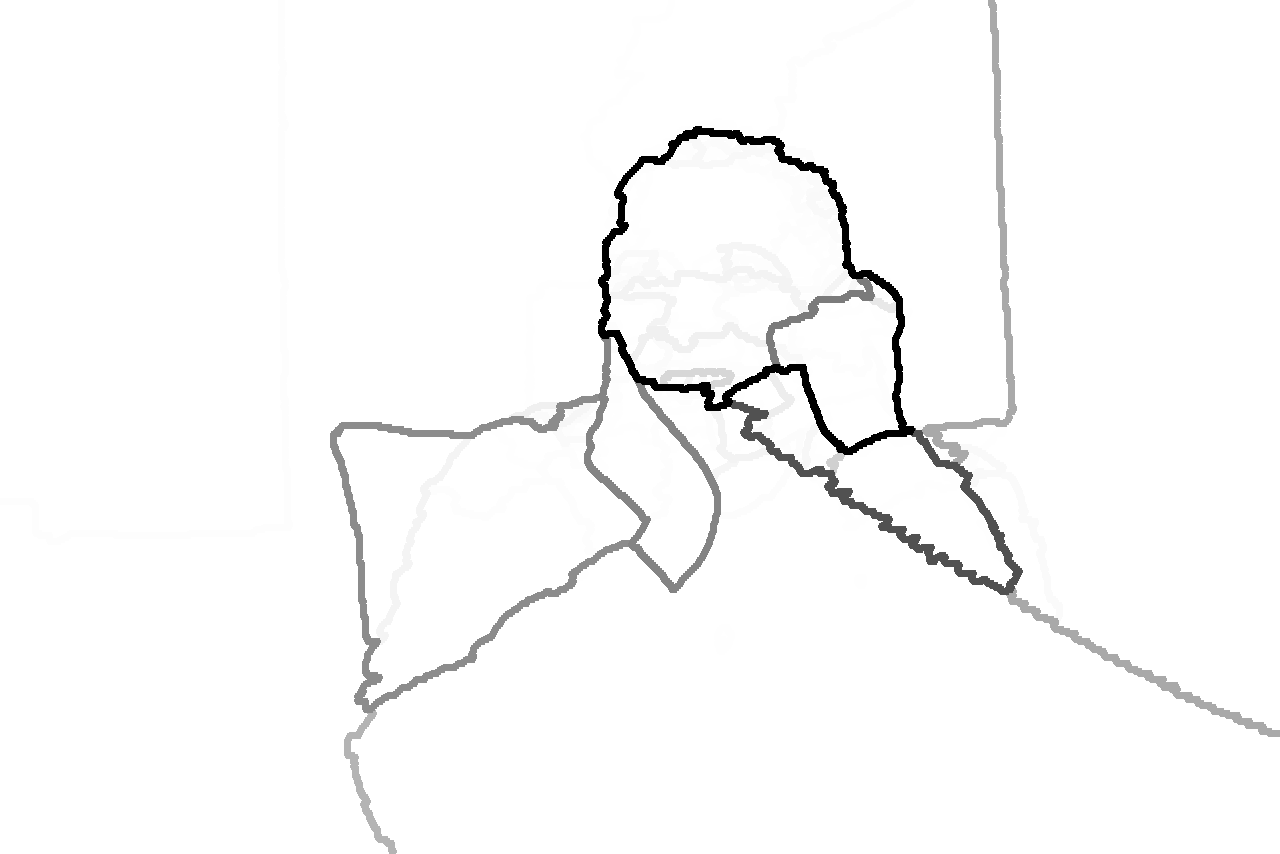}} \\
\squeezeup
\subfigure[]{\includegraphics[width=.325\columnwidth]{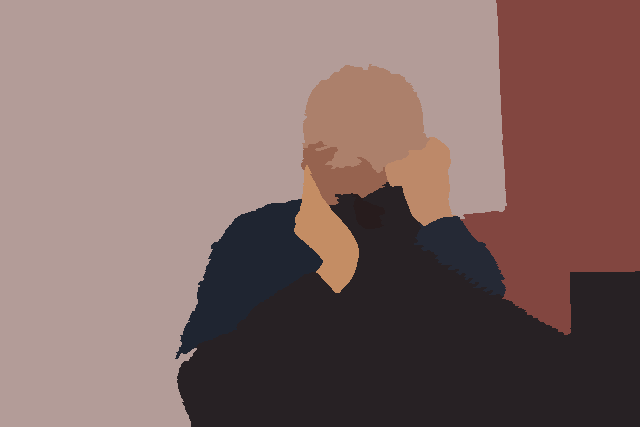}}
\subfigure[]{\includegraphics[width=.325\columnwidth]{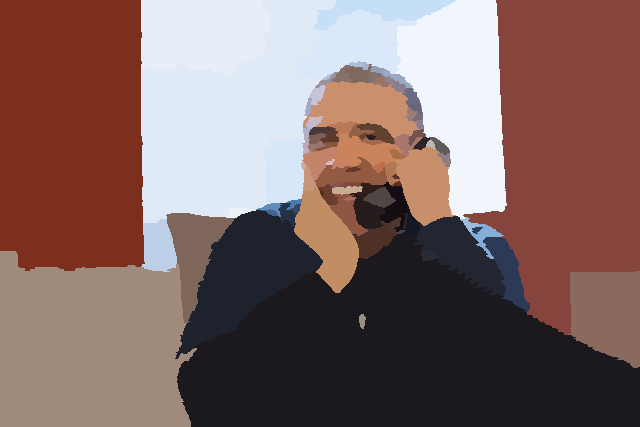}}
\subfigure[]{\includegraphics[width=.325\columnwidth]{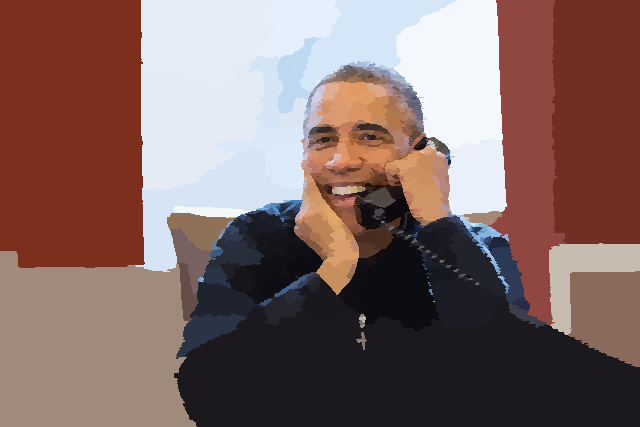}}
\squeezeup
\squeezeup
\caption{Hierarchical segmentation of faces. (a) Original image (b) Prior : Probabilistic map obtained thanks to face detection algorithm and (c) Volume-based SWS Hierarchy UCM (d) Volume-based HRF UCM with face position as prior (e)(f)(g) Examples of segmentations obtained with HRF - 10,100,1000 regions. }
\label{fig:FaceHRF}
\end{center}
\end{figure}

\squeezeup
\squeezeup
\squeezeup
\squeezeup

\section{Application examples}
\label{sec:examples}
\subsection{Scalable transmission favoring regions of interest}
Let us consider a situation where one emitter wants to transmit an image through a channel with a limited bandwidth, e.g. for a videoconference call. In such a case, the more important informations to transmit are details on the face of the person on the image. Besides, we nowadays have highly efficient face detectors, using for example Haar-wavelets as features in a learning-based vision approach \cite{viola01}. Considering that for an image in entry, the face can be easily detected, we can use this information to produce a hierarchical segmentation of the image that accentuates the details around the face while giving a good sketch of the image elsewhere. Depending on the bandwidth available, we can then choose the level of the hierarchy to select and obtain the associated partition to transmit, ensuring us to convey the face with as much details as possible. Some results are presented in Figure \ref{fig:FaceHRF}, with notably a comparison between a classical volume-based SWS UCM and a volume-based HRF UCM.

\begin{figure}[H]
\squeezeup
\begin{center}
\subfigure[]{\includegraphics[width=.242\columnwidth]{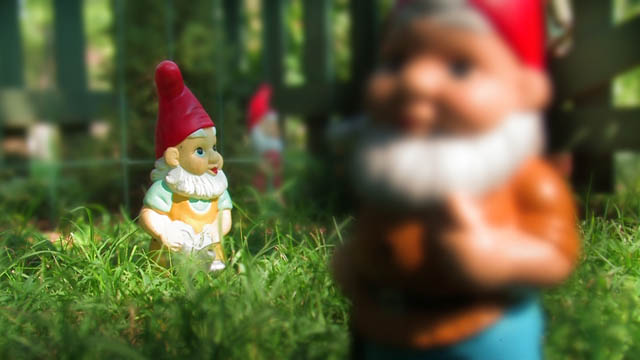}}
\subfigure[]{\includegraphics[width=.242\columnwidth]{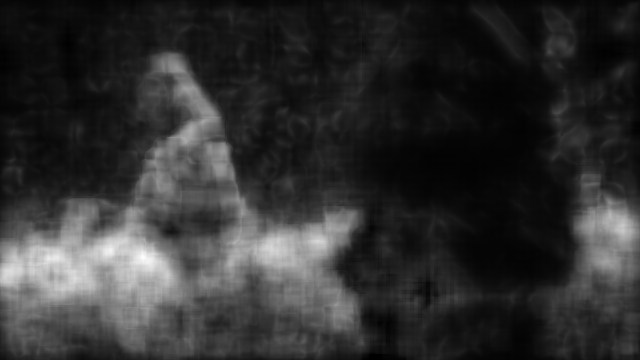}}
\subfigure[]{\includegraphics[width=.242\columnwidth]{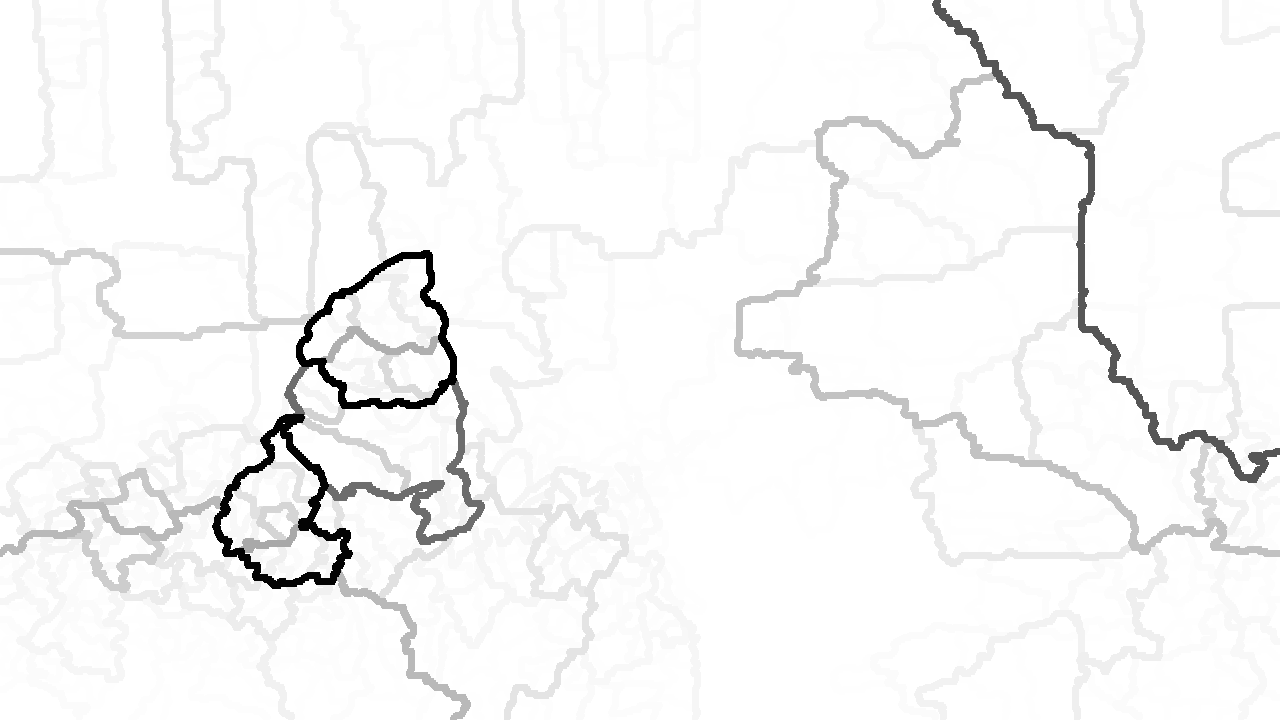}}
\subfigure[]{\includegraphics[width=.242\columnwidth]{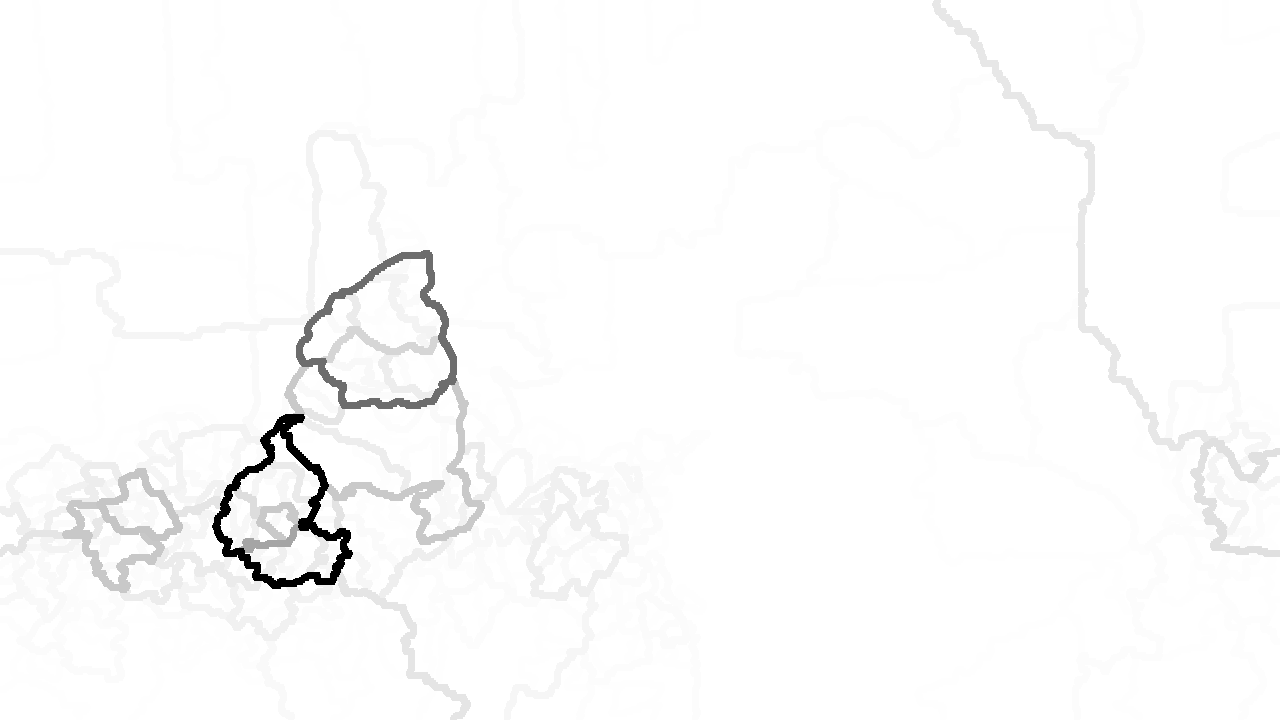}} \\
\squeezeup
\subfigure[]{\includegraphics[width=.325\columnwidth]{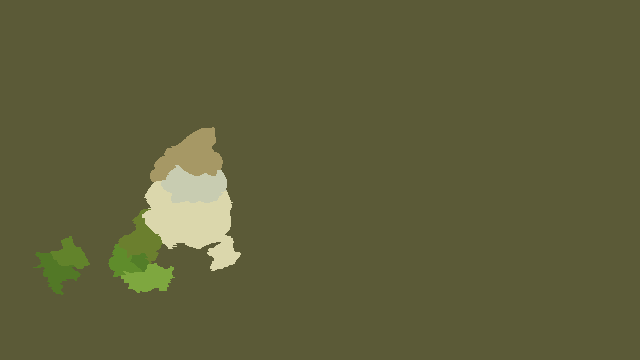}}
\subfigure[]{\includegraphics[width=.325\columnwidth]{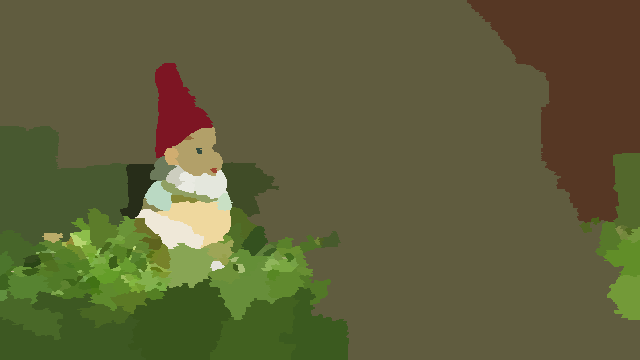}}
\subfigure[]{\includegraphics[width=.325\columnwidth]{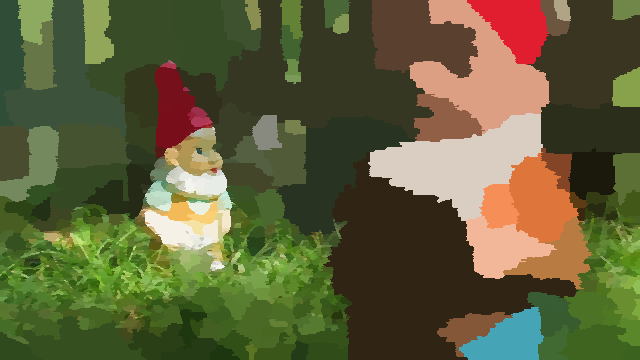}}
\squeezeup
\squeezeup
\caption{Hierarchical segmentation of non-blur objects. (a) Original image (b) Prior image obtained with non-blur zones detection algorithm  (c) Volume-based SWS Hierarchy UCM (d) Volume-based HRF UCM with non-blur zones as prior (e)(f)(g) Examples of segmentations obtained with HRF - 10,200,2000 regions.}\label{Fig:blur}
\end{center}
\end{figure}

\squeezeup
\squeezeup
\squeezeup
\squeezeup
\squeezeup
\squeezeup
\subsection{Artistic aspect: focus and cartoon effect}
The same method can also be used for artistic purposes. For example, when taking as prior the result of a blur detector \cite{su2011}, we can accentuate the focus effect wanted by the photograph and turn it into a cartoon effect as well - see Figure \ref{Fig:blur} for an illustration of the results. 

\begin{figure}[H]
\squeezeup
\squeezeup
\squeezeup
\begin{center}
\subfigure[]{\includegraphics[width=.225\columnwidth]{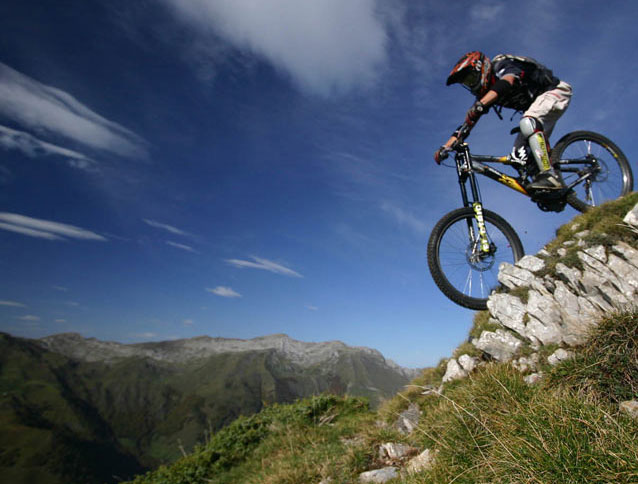}}
\subfigure[]{\includegraphics[width=.225\columnwidth]{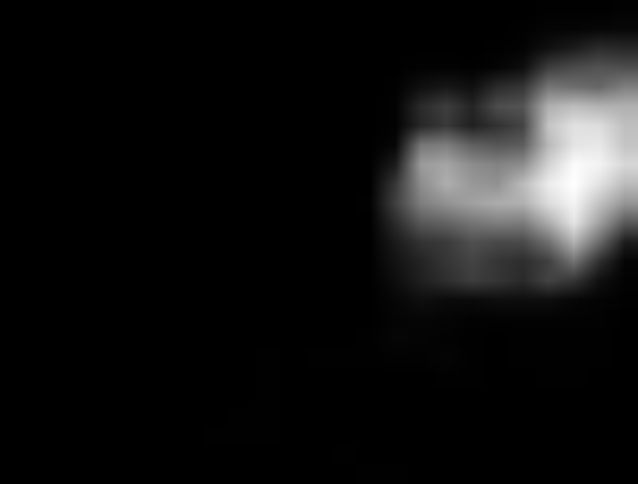}}
\subfigure[]{\includegraphics[width=.225\columnwidth]{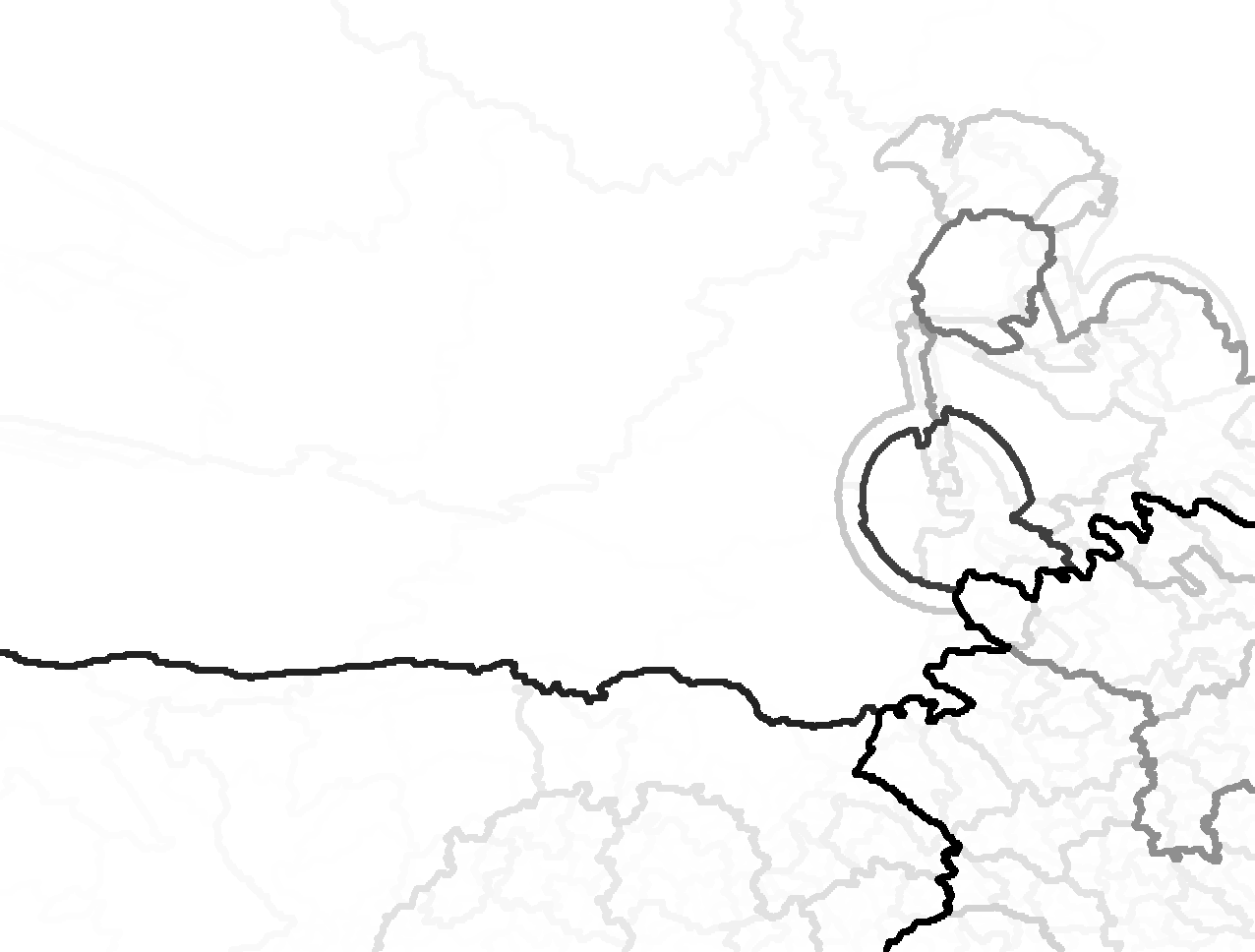}}
\subfigure[]{\includegraphics[width=.225\columnwidth]{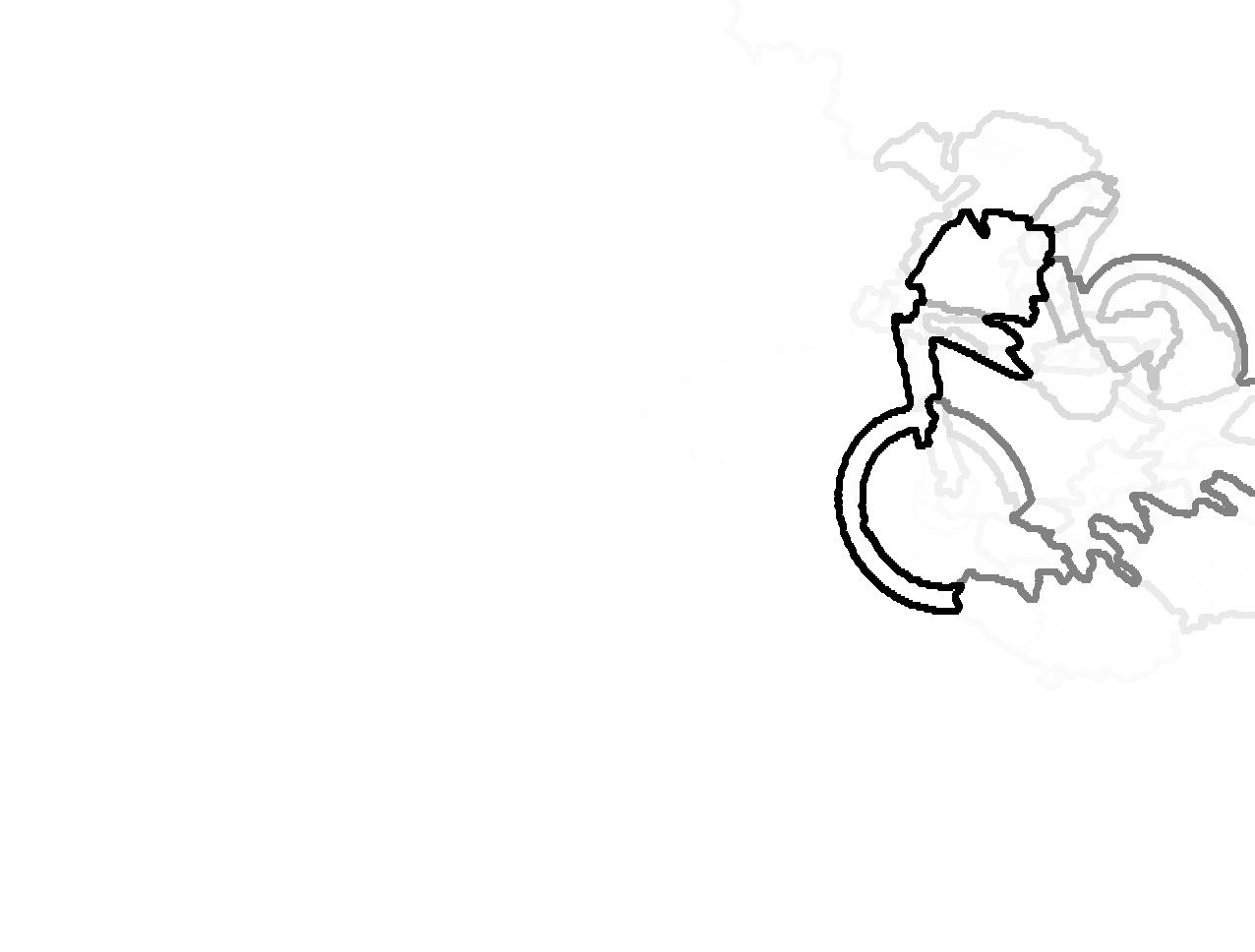}} \\
\squeezeup
\subfigure[]{\includegraphics[width=.325\columnwidth]{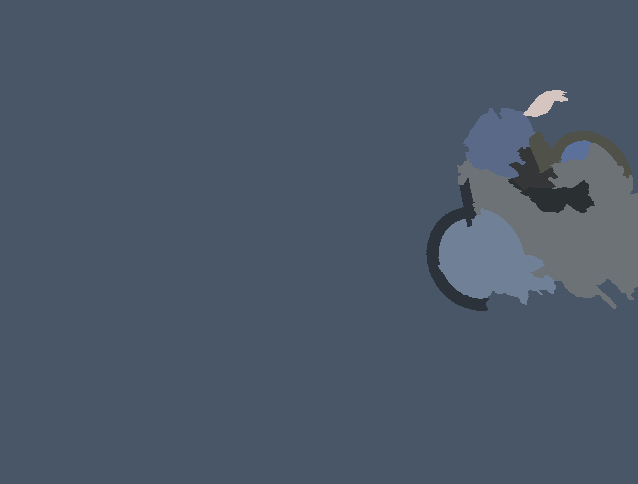}}
\subfigure[]{\includegraphics[width=.325\columnwidth]{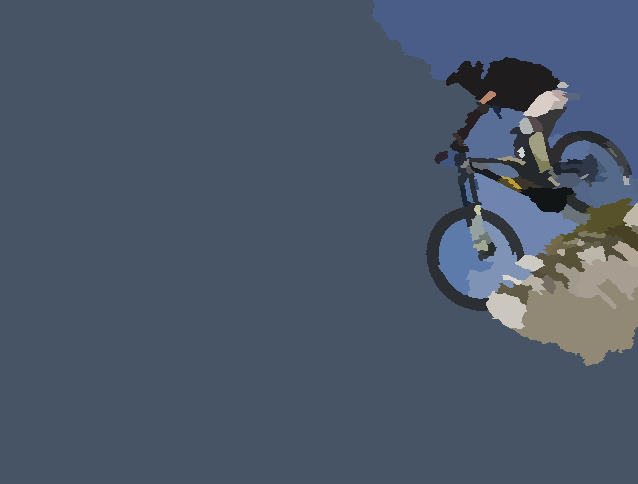}}
\subfigure[]{\includegraphics[width=.325\columnwidth]{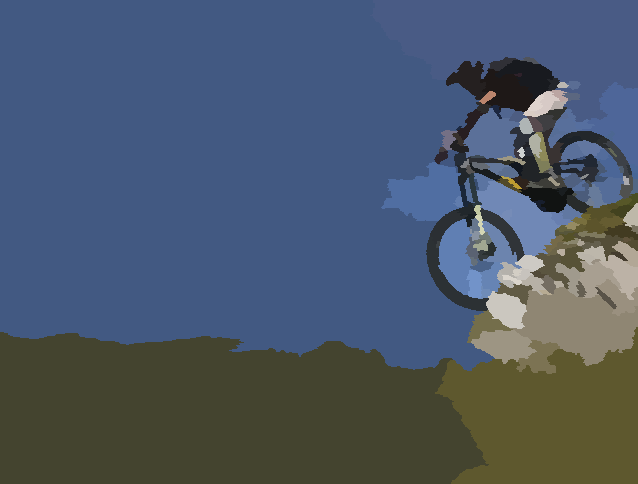}}
\squeezeup
\squeezeup
\caption{Hierarchical segmentation of the main class (class ''bike") in an image with heatmap issued of a CNN-based method as input. (a) Original image, (b) Heatmap issued by the CNN-based localization method, (c) Volume-based Watershed Hierarchy UCM and (d) Hierarchy with prior UCM. (e)(f)(g) Examples of segmentations obtained with HRF - 10,100,200 regions.}\label{Fig:CNN}
\end{center}
\end{figure}

\begin{figure}[H]
\squeezeup
\squeezeup
\squeezeup
\squeezeup
\squeezeup
\squeezeup

\begin{center}
\subfigure[]{\includegraphics[width=.18\columnwidth]{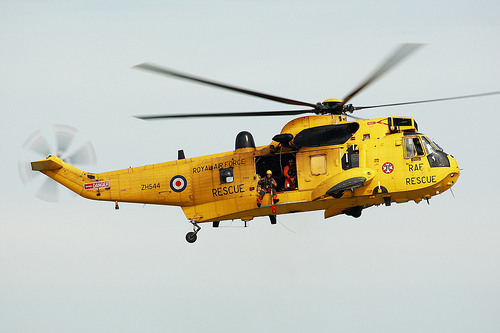}}
\subfigure[]{\includegraphics[width=.18\columnwidth]{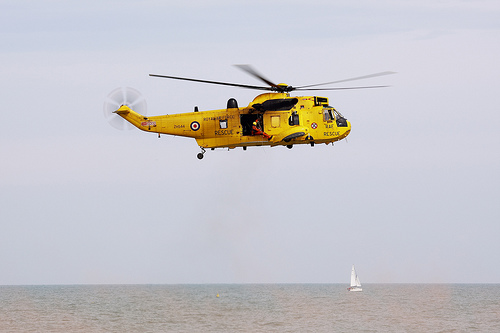}}
\subfigure[]{\includegraphics[width=.12\columnwidth]{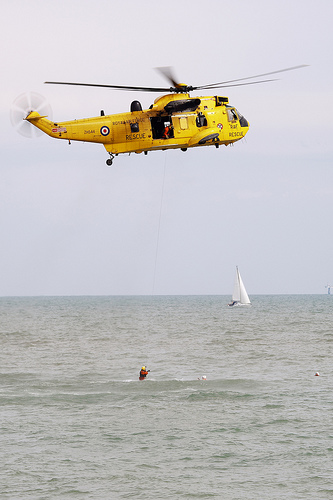}}
\subfigure[]{\includegraphics[width=.12\columnwidth]{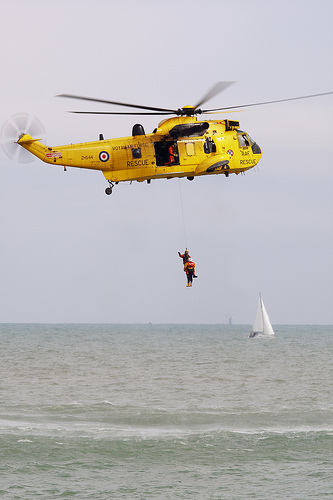}}
\subfigure[]{\includegraphics[width=.18\columnwidth]{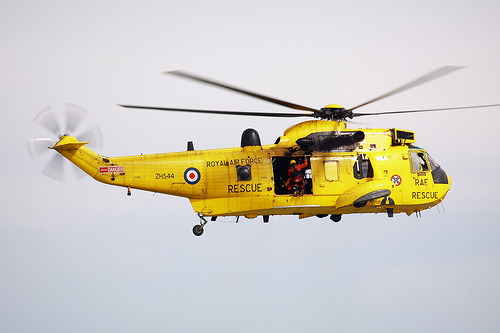}}\\
\squeezeup
\subfigure[]{\includegraphics[width=.15\columnwidth]{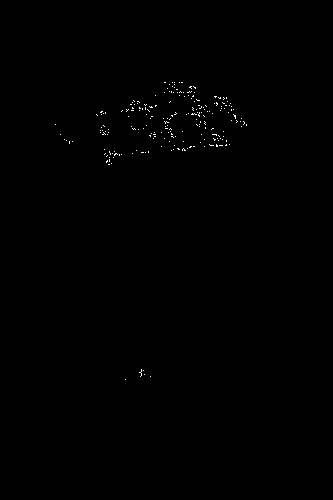}}
\subfigure[]{\includegraphics[width=.15\columnwidth]{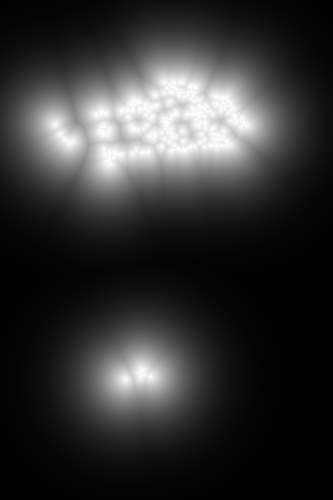}}
\subfigure[]{\includegraphics[width=.15\columnwidth]{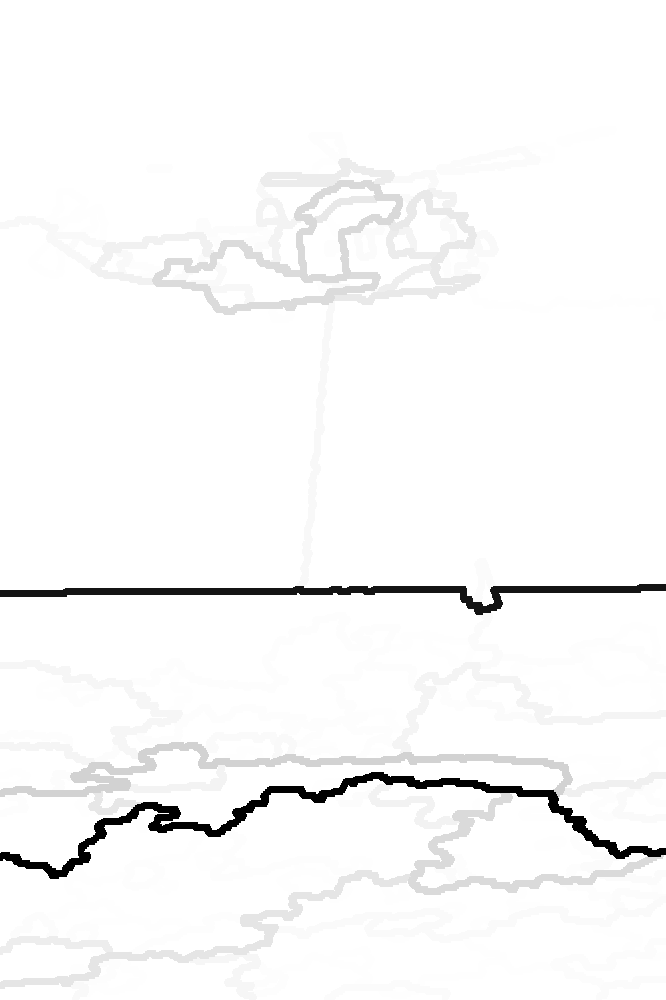}}
\subfigure[]{\includegraphics[width=.15\columnwidth]{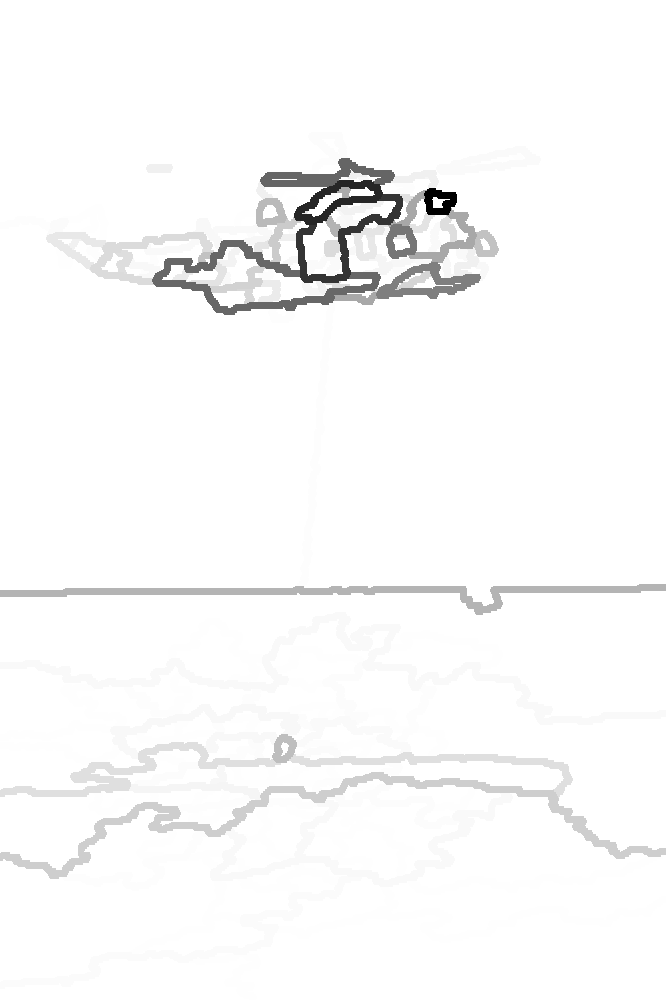}}
\subfigure[]{\includegraphics[width=.15\columnwidth]{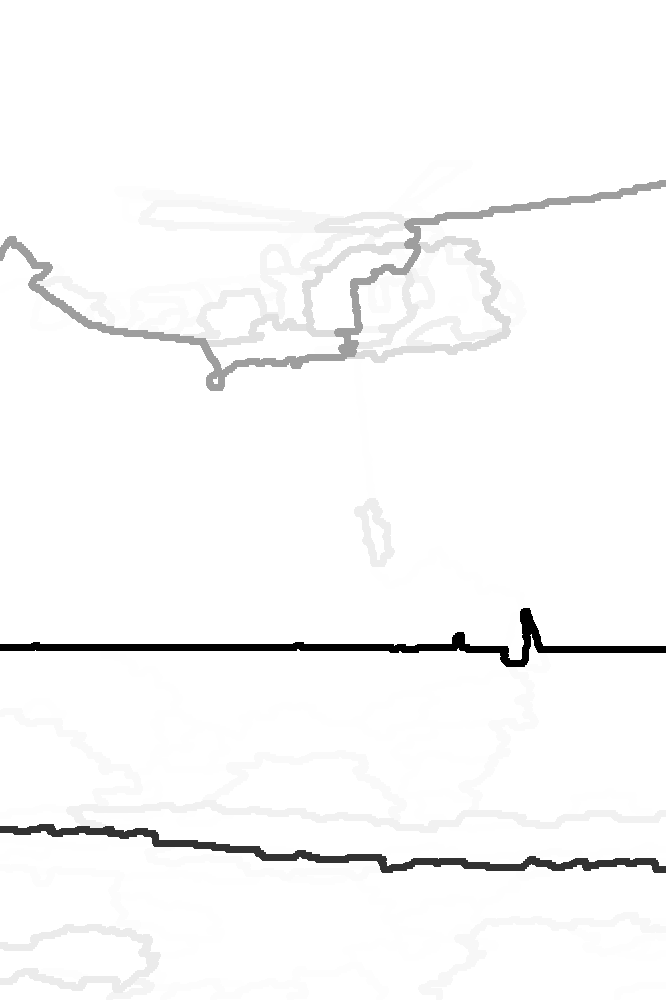}}
\subfigure[]{\includegraphics[width=.15\columnwidth]{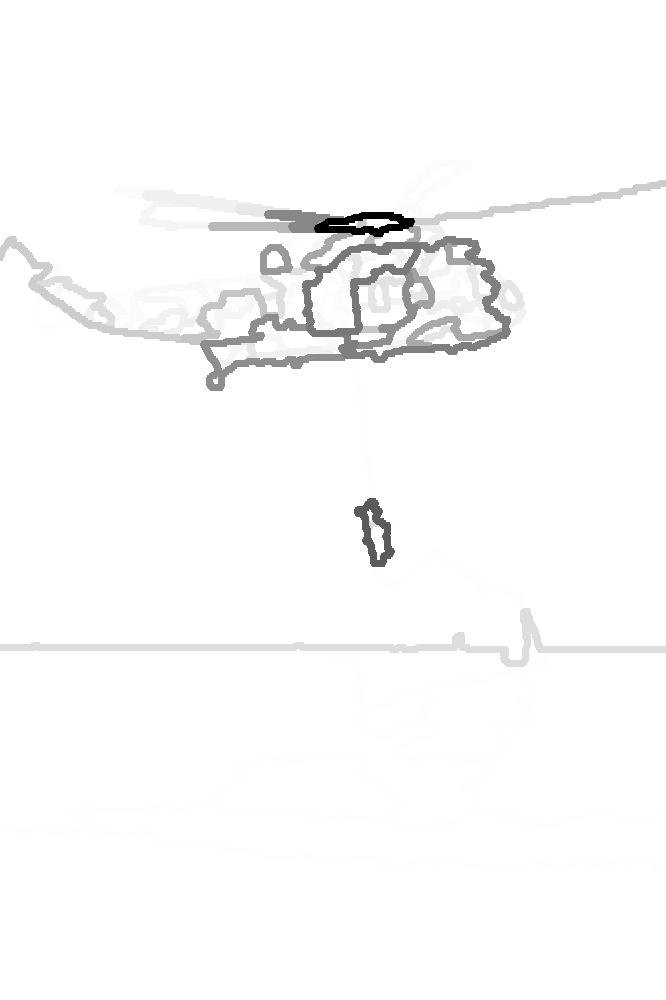}}
\squeezeup
\squeezeup
\caption{Hierarchical co-segmentation of matched objects. (a)-(e) Images $I_{1}$ to $I_{5}$ (f) All matched key-points of $I_{3}$ with other images key-points (g) Prior for $I_{3}$: probability map generated thanks to morphological distance function (h)(i) Volume-based SWS Hierarchy UCM for $I_{3}$ and $I_{4}$ (j)(k) Volume-based HRF UCM for $I_{3}$ and $I_{4}$ with matched key-points as prior.}\label{Fig:Coseg}
\end{center}
\end{figure}

\squeezeup
\squeezeup
\squeezeup
\squeezeup
In the same spirit, various methods now exist to automatically roughly localize the principal object in an image. We inspire ourselves from \cite{oquab15} to do so. Using the state-of-the-art convolution neural network (CNN) classifier VGG19 \cite{simonyan14} trained on the 1000 classes ImageNet database \cite{deng09}, we first determine what is the main class in the image. Note that this CNN takes as input only images of size $224\times224$ pixels. Once it is known, we can then, by rescaling the image by a factor $s \in \{0.5,0.7,1.0,1.4,2.0,2.8\}$, compute for sub-windows of size $224\times224$ of the image the probability of appearance of the main class. By simply superimposing the results for all sub-windows, we thus obtain a probabilistic map of the main class for each rescaling factor. By max-pooling, we keep in memory the result of the scale for which the probability is the highest. The \textit{heatmap} thus produced can then be used to feed our algorithm. This way, we have at our disposal an automatized way to focus on the principal class in the scene. Some results are presented in Figure \ref{Fig:CNN}.

\begin{figure}
\squeezeup
\squeezeup
\squeezeup
\begin{center}
\subfigure[]{\includegraphics[width=.242\columnwidth]{./images/Face/Second/imIn}}
\subfigure[]{\includegraphics[width=.242\columnwidth]{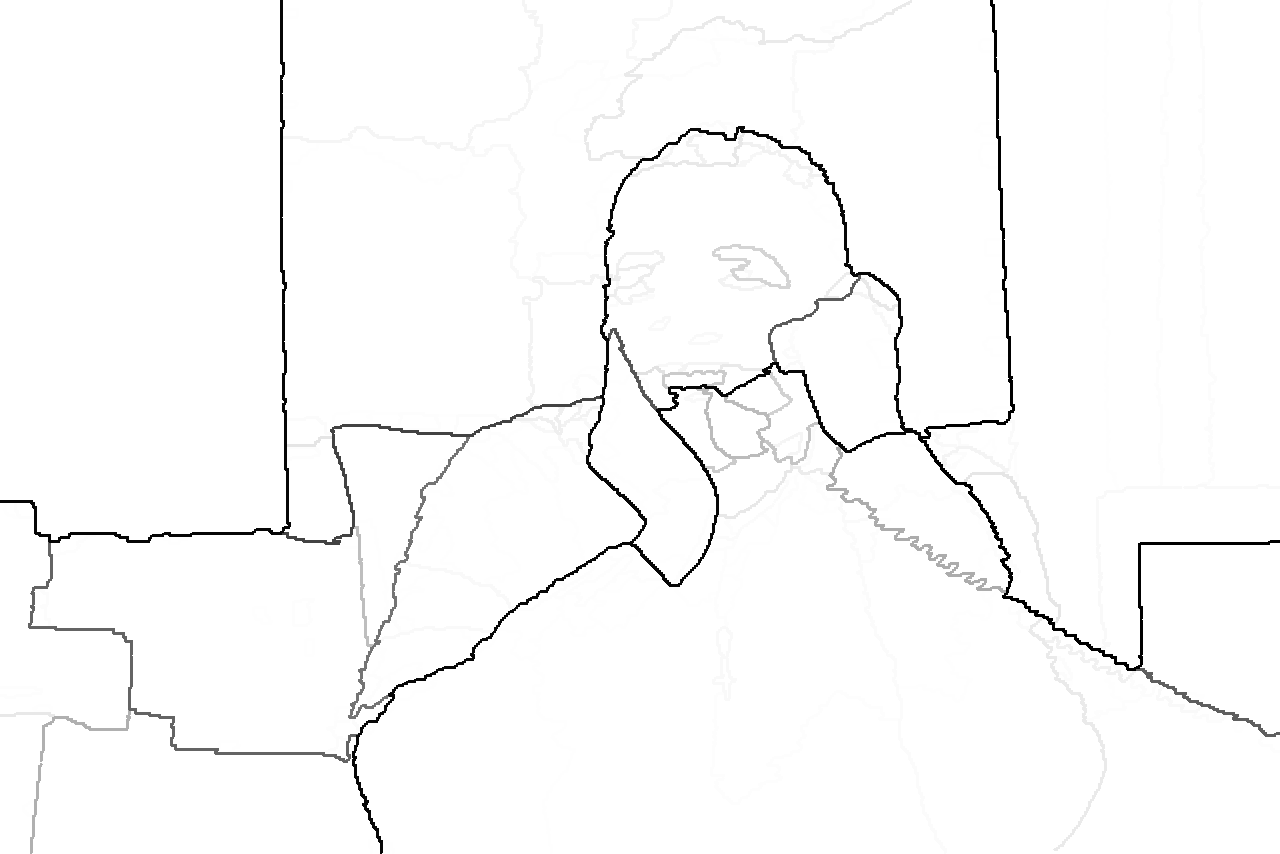}}
\subfigure[]{\includegraphics[width=.242\columnwidth]{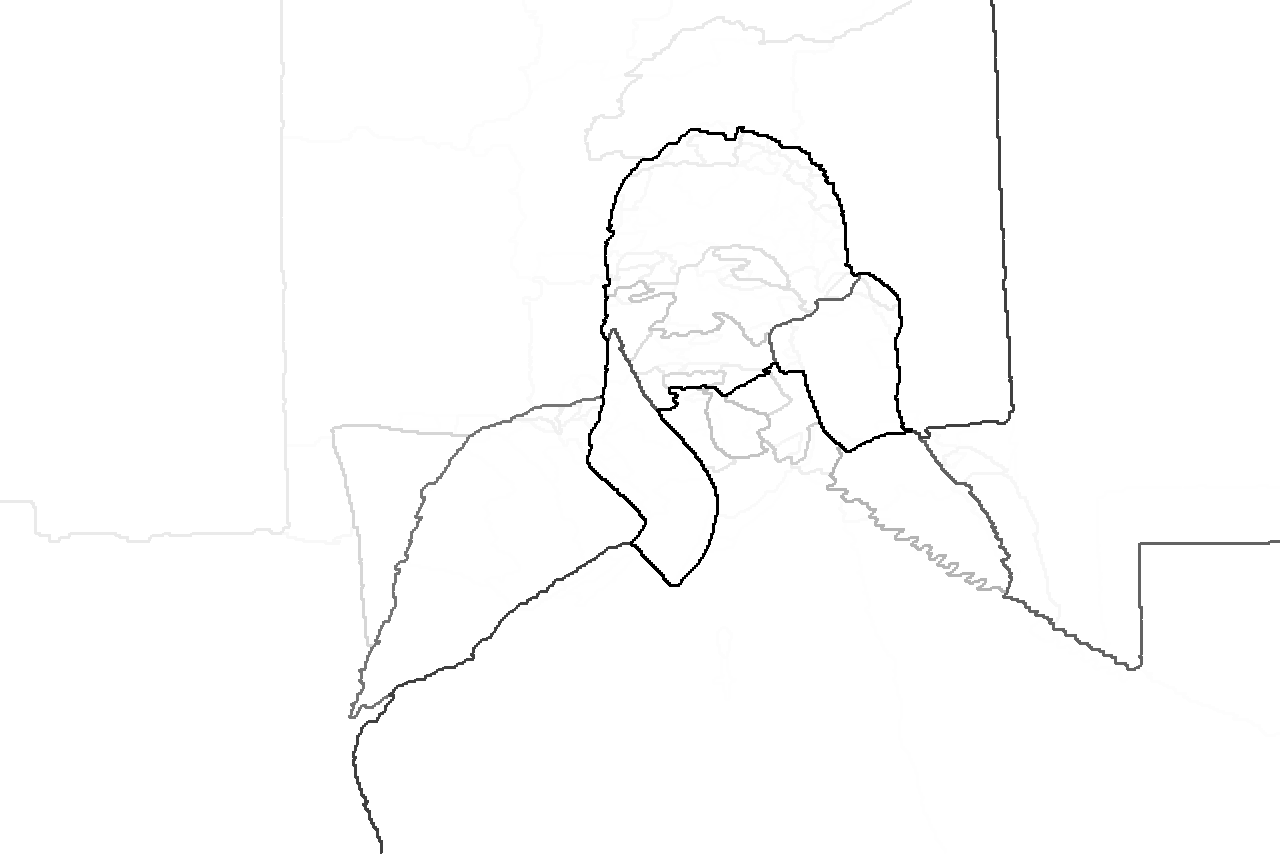}}
\subfigure[]{\includegraphics[width=.242\columnwidth]{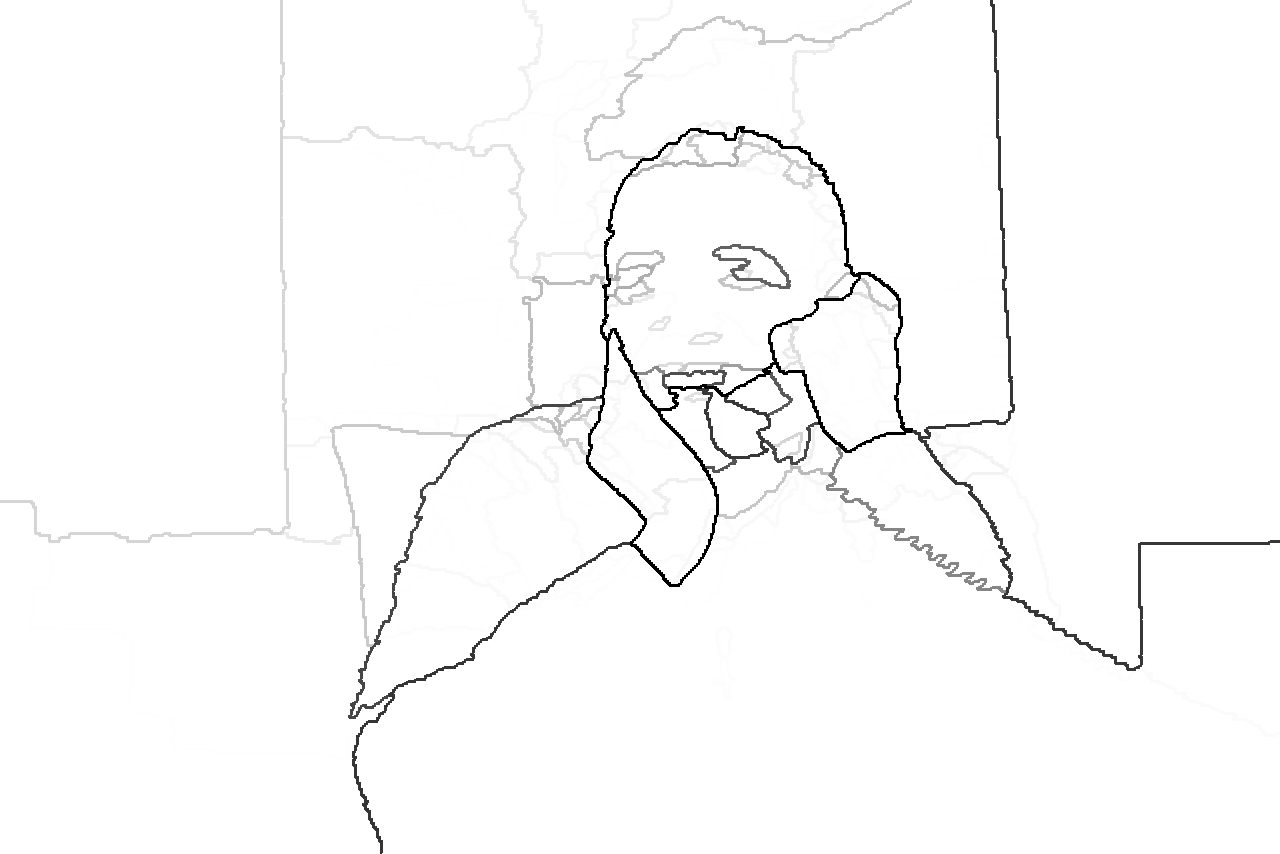}} \\
\squeezeup
\subfigure[]{\includegraphics[width=.325\columnwidth]{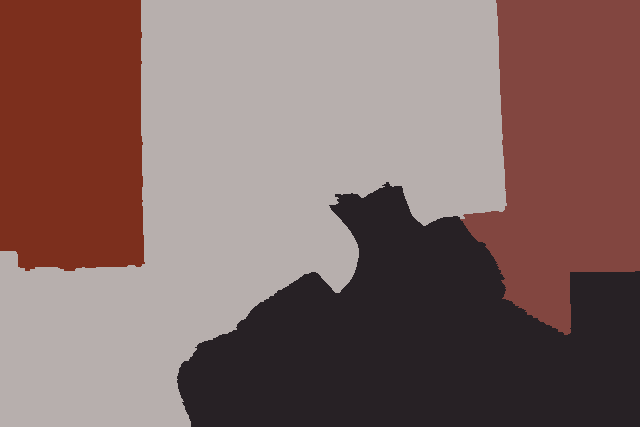}}
\subfigure[]{\includegraphics[width=.325\columnwidth]{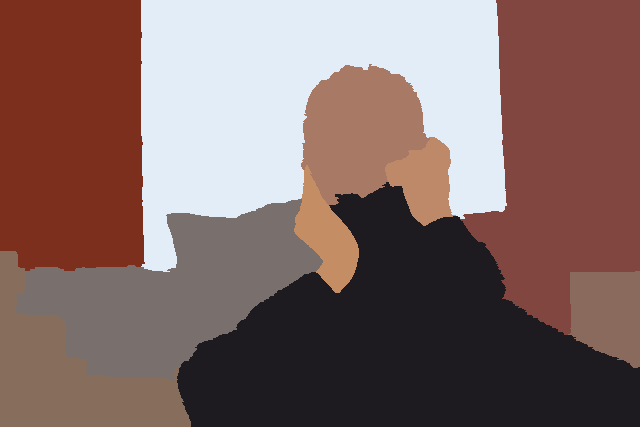}}
\subfigure[]{\includegraphics[width=.325\columnwidth]{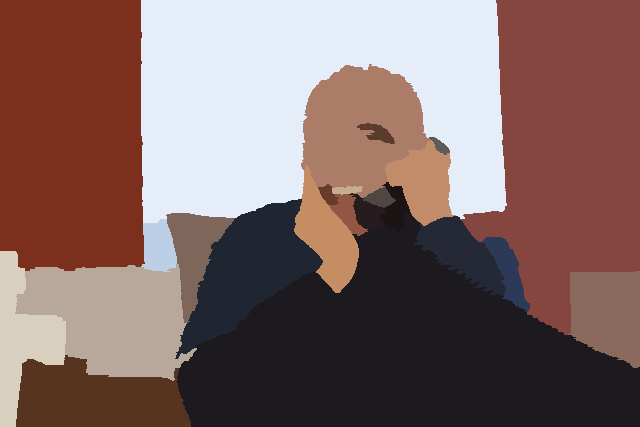}} \\
\squeezeup
\subfigure[]{\includegraphics[width=.325\columnwidth]{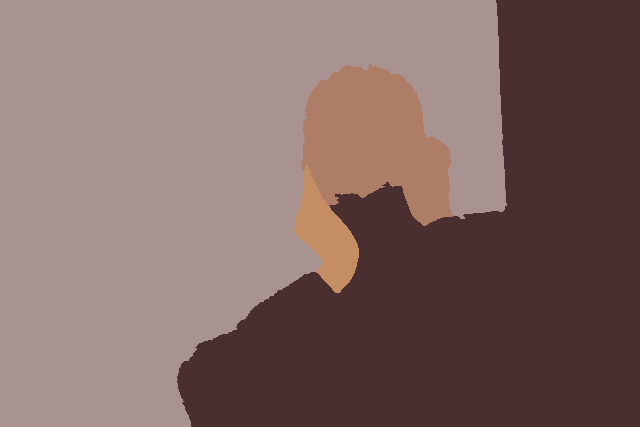}}
\subfigure[]{\includegraphics[width=.325\columnwidth]{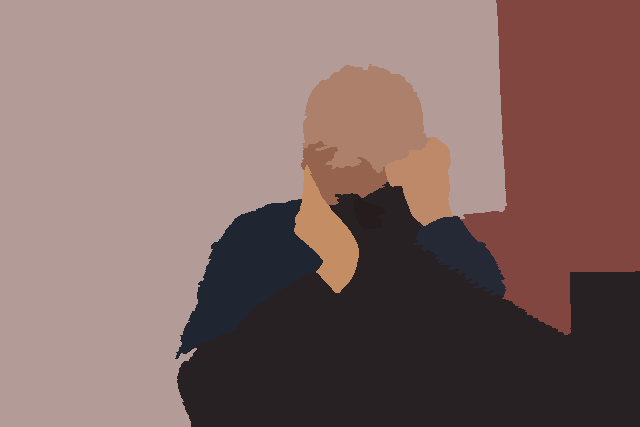}}
\subfigure[]{\includegraphics[width=.325\columnwidth]{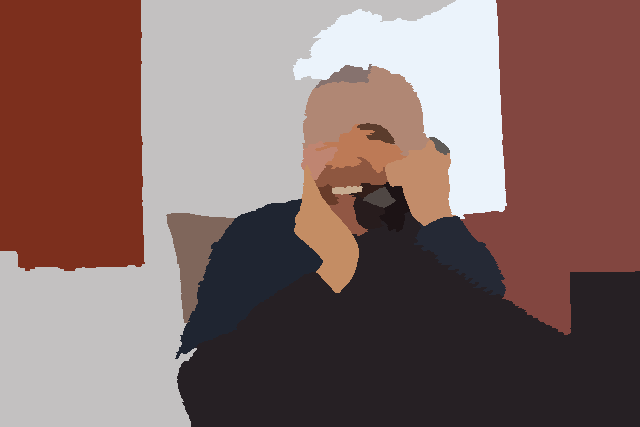}} \\
\squeezeup
\subfigure[]{\includegraphics[width=.325\columnwidth]{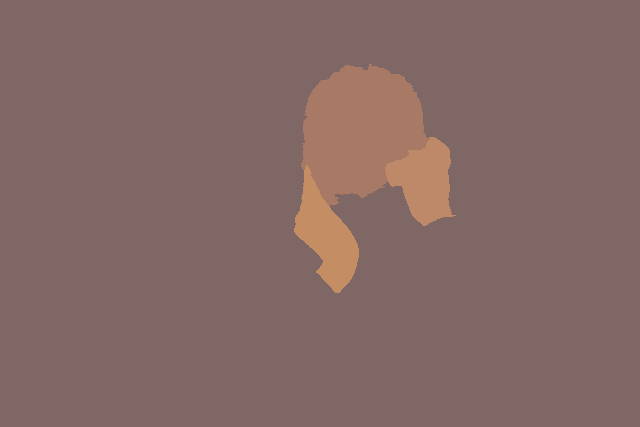}}
\subfigure[]{\includegraphics[width=.325\columnwidth]{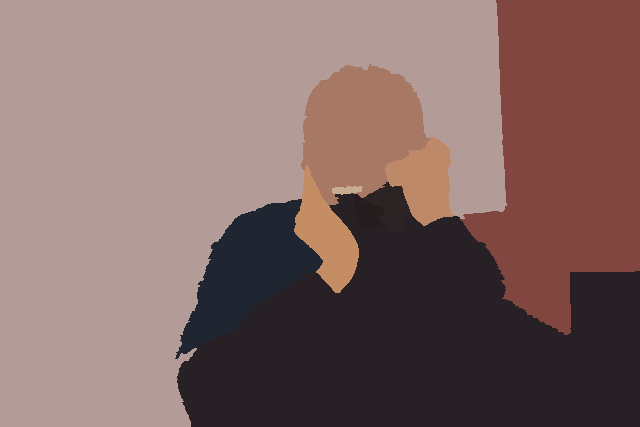}}
\subfigure[]{\includegraphics[width=.325\columnwidth]{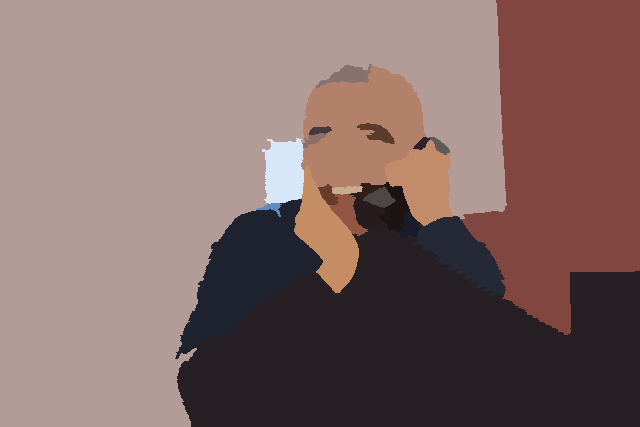}}
\squeezeup
\squeezeup
\caption{Hierarchical segmentation of faces. (a) Original image (b)(c)(d) UCM of HRF depending of the couples of regions (section \ref{ssec:varHRF}), of the regions only (section \ref{ssec:HRF}), and both combined. The rest of the images are segmentations examples with 4,10,25 regions, the hierarchies being presented in the same order.}
\label{fig:FaceHRFbis}
\end{center}
\end{figure}

\squeezeup
\squeezeup
\squeezeup
\squeezeup
\squeezeup
\squeezeup
\subsection{Hierarchical co-segmentation}
Another potential application is to co-segment with the same fineness level an object appearing in several different images. For example, when given a list of images of the same object taken from different perspectives/for different conditions, we can follow the state-of-the-art matching procedure \cite{lowe04}: (i) compute all key-points in both images, (ii) compute local descriptors at these key-points, (iii) match those key-points using a spatial coherency algorithm as RANSAC. Once it is done we retain these matched key-points for both images, and generate probability maps of the appearance of the matched objects using a morphological distance function to the matched key-points.

These probability maps can then feed our algorithm, given as result a hierarchical co-segmentation that emphasizes the matched zones of the image. Some results are presented in Figure \ref{Fig:Coseg}.

\subsection{Example of the effect of the HRF highlighting transitions between foreground and background}
We illustrate here the HRF highlighting transitions between foreground and background presented in section \ref{ssec:varHRF}, by presenting its effect in the face detection example presented in figure \ref{fig:FaceHRFbis}.
 
\section{Conclusions and perspectives}
\label{sec:conclusion}
In this paper we have proposed a novel and efficient hierarchical segmentation algorithm that emphasizes the regions of interest in the image by using spatial exogenous information on it. The wide variety of sources for this exogenous information makes our method extremely versatile and its potential applications numerous, as shown by the examples developed in the last section. To go further, we could find a way to efficiently extend this work to videos. One could also imagine a semantic segmentation method that would go back and forth between localization algorithm and HRF to progressively refine the contours of the main objects in the image.

\FloatBarrier
\bibliographystyle{splncs03}
\bibliography{SegmentationBib}
\end{document}